\documentclass[10pt]{cai26}
\usepackage{microtype}
\usepackage{graphicx}
\usepackage{subcaption}
\usepackage{booktabs} 
\usepackage{algorithm}
\usepackage{algpseudocode}

\begin{document}
\def\conferenceyear{2026}
\volumeheader{39}{0}
\begin{center}

\title{Are You the A-hole?\\A Fair, Multi-Perspective Ethical Reasoning Framework}
\maketitle

\thispagestyle{empty}
\pagenumbering{gobble}

\begin{tabular}{c}
Sheza Munir\upstairs{*}, Ahanaf Rodoshi, Sumin Lee, \\
Feiran Chang, Xujie Si, Syed Ishtiaque Ahmed \\
[0.5ex]
{\small Computer Science, University of Toronto}
\end{tabular}
  
\emails{
  \upstairs{*} corresponding author: sheza@cs.toronto.edu
}
\vspace*{0.1in}

\begin{abstract}
Standard methods for aggregating natural language judgments, such as majority voting, often fail to produce logically consistent results when applied to high-conflict domains, treating differing opinions as noise. We propose a neuro-symbolic aggregation framework that formalizes conflict resolution through Weighted Maximum Satisfiability (MaxSAT). Our pipeline utilizes a language model to map unstructured natural language explanations into interpretable logical predicates and confidence weights. These components are then encoded as soft constraints within the Z3 solver, transforming the aggregation problem into an optimization task that seeks the maximum consistency across conflicting testimony.\footnote{Code and data: \url{https://github.com/ShezaMunir/Ethical-Reasoning-Framework.git}} Using the Reddit \texttt{r/AmItheAsshole} forum as a case study in large-scale moral disagreement, our system generates logically coherent verdicts that diverge from popularity-based labels 62\% of the time, corroborated by an 86\% agreement rate with independent human evaluators. This study demonstrates the efficacy of coupling neural semantic extraction with formal solvers to enforce logical soundness and explainability in the aggregation of noisy human reasoning. 
\end{abstract}
\end{center}

\copyrightnotice


\section{Introduction}

The aggregation of human judgments serves as the bedrock of modern artificial intelligence; however, the reliance on majority voting schemes in data annotation has long served as a systemic source of epistemic injustice. By definition, majority-based aggregation techniques ensure that minority perspectives (often those most vulnerable or distinct from the cultural norm) are statistically drowned out by the dominant viewpoint. As AI systems increasingly engage with controversial, ambiguous, and high-stakes real-world situations, ranging from ethical classification to automated social judgment, the limitations of simple consensus become dangerously apparent. There is an urgent, growing need for principled methods to aggregate diverse human opinions that go beyond mere vote-counting. Unlike less subjective domains such as object recognition, where a ``ground truth" (e.g., \textit{``this is a cat"}) is indisputable, moral judgments are inherently pluralistic, contested, and deeply context-dependent. Yet, the vast majority of current crowdsourcing and annotation approaches continue to rely on consensus-based labeling strategies. These methods risk oversimplifying complex ethical reasoning, flattening cultural variance, and amplifying biases rooted in popularity, temporal availability, or herd behavior.

This persistent reliance on consensus is a root cause of bias and unfairness in deployment-ready AI models. When constructing ``ground truth" datasets for subjective tasks, the reasoning underlying a label, the \textit{“why”}, is arguably just as critical, if not more so, than the annotation decision, the \textit{“what.”} For instance, an individual’s lived experience in a high-conflict geopolitical environment may lead to a radically different, yet equally valid, interpretation of hate speech compared to an annotator unfamiliar with those specific social norms or historic tensions. Current aggregation methods fail to capture this nuance; by allowing the majority vote to act as the sole arbiter of truth, these systems effectively create epistemic injustice, suppressing valid minority perspectives in favor of ``hivemind" behavior or emotional bias. In such paradigms, disagreement is treated as noise to be eliminated rather than a signal to be analyzed. Therefore, achieving truly fair aggregation requires a fundamental paradigm shift: moving the goalpost from collecting votes to aggregating \textit{reasoning}, prioritizing the justification provided for a label rather than the raw numerical count.

A distinct and rich illustration of these aggregation failures can be observed in the subreddit \texttt{r/AmItheAsshole} (AITA), a massive online community where users post detailed moral conundrums for public judgment. Commenters assign standardized labels such as \textsc{NTA} (Not the Asshole), \textsc{YTA} (You're the Asshole), \textsc{NAH} (No Assholes Here), or \textsc{ESH} (Everyone Sucks Here), accompanied by their reasoning for why they believe so. Currently, the platform employs a heuristic ``Judgement Bot" that locks the decision after a fixed window of 18 hours, basing the final verdict solely on the single most upvoted comment. This approach, representative of common moderation techniques on social media, is fraught with structural unfairness and distinct biases: it ignores critical details or counter-arguments added later in the discussion; it suffers from geographical disconnects (e.g., North American users judging Asian cultural contexts simply because of posting time zones); and it effectively reduces complex morality to a popularity contest where minority voices are structurally excluded. The mechanical imposition of a time limit combined with a ``winner-takes-all" voting system perfectly encapsulates the flaws of democratic consensus applied to nuanced ethics.

To address these profound limitations, we propose a novel, reasoning-centered framework for aggregating multi-annotator moral judgments that explicitly models ethical predicates and constraints. Rather than treating biased, minority, or dissenting comments as statistical outliers to be discarded, our framework treats each comment as a valuable unit of reasoning composed of identifiable predicates, such as \textit{harm}, \textit{intentionality}, \textit{negligence}, or \textit{apology}. Using AITA as a focus study, we extract these logical reasoning patterns from natural language comments and translate them into formal symbolic representations. A constraint solver, Z3 \cite{z3solver}, is then utilized to determine which conclusions are logically consistent across the diverse set of perspectives. This process produces a final label grounded in logical \textit{reasoning} rather than mere \textit{consensus}, ensuring that the final verdict respects the internal logic of ethical argumentation rather than just the volume of votes.

\subsection{Contributions}
This study makes the following key contributions:

\begin{itemize}
    \item \textbf{A "split-stream" architecture for ethical reasoning:} We introduce a novel aggregation pipeline that architecturally decouples factual claims (the \textit{logic stream}) from character judgments (the \textit{ethic stream}). By weighting these streams independently based on argumentation quality rather than annotator quantity, our system computationally enforces the principle that epistemic authority and sound reasoning should outweigh democratic popularity in moral decision-making processes.
    
    \item \textbf{Formalization of jurisprudence into logic constraints:} We bridge the gap between abstract moral philosophy and computational verification by translating established frameworks into quantifiable soft constraints. This formalization allows a Z3 solver to derive verdicts that are ethically coherent and logically sound, rather than just statistically probable.
    
    \item \textbf{Empirical evidence of a "strictness shift":} Through a rigorous analysis of 600 high-conflict scenarios, we demonstrate that prioritizing high-quality reasoning over majority voting leads to a measurable increase in accountability. Our system overturned 62\% of majority-vote decided verdicts, leaning more towards higher accountability. This validates the hypothesis that reasoning-based aggregation can pierce through the "echo chamber" of online validation, revealing a stricter, more principled moral ground truth hidden beneath the noise of the crowd.
\end{itemize}

\section{Background}

The AI ethics domain posits that accountability necessitates human moral reasoning beyond simple numerical metrics. This study develops a reasoning-based aggregation framework for Reddit r/AmItheAsshole posts, treating each comment as a moral annotation. Using LLM-based feature extraction, logical rules, and solver-based verification, we formalize these judgments and derive a final label through ethical coherence rather than the existing majority-vote approach.

Zhang et al.'s MAGI (Multi-Annotated Explanation-Guided Learning) models diverse human explanations as probabilistic latent variables rather than treating disagreement as noise, generating predictions more aligned with human reasoning \citep{zhang2023magi}. However, it cannot explain moral decisions beyond latent factors or ensure ethical coherence. Similarly, Gordon et al.'s Jury Learning aggregates judgments from selected ``moral juries'' to make value choices transparent, but the aggregation remains statistical and lacks guarantees of logical consistency or normative alignment\citep{gordon2022jurylearning}. Both approaches lack two components central to our pipeline: LLM-based extraction of structured moral features and formal verification via Z3, which ensures coherent and contradiction-free aggregation.

The Reddit AITA forum is an appropriate and well-motivated choice for studying moral decision-making because it contains real-life moral dilemmas paired with community judgments, which makes it one of the richest natural sources of descriptive moral reasoning. Sabri and Bsher show that AITA provides diverse, high-quality moral labels that reflect genuine social norms \citep{bsher2023aita}. Likewise, Lourie et al. assert AITA as a rare large-scale dataset where people organically debate ethical responsibility, offering the distributional judgments needed to study disagreement and controversy \citep{lourie2021scruples}.

We use LLM reasoning in our pipeline to extract the moral features from the AITA thread comments. Chakraborty et al. show that structured moral prompting by using explicit ethical principles such as first-person reasoning and care ethics, raises LLM coherence by 18\% \citep{chakraborty2025structured}.This indicates that LLMs reason more reliably when their decisions are scaffolded by moral structure. Motivated by this, our prompt engineering explicitly foregrounded ethical principles within the LLM’s instructions. However, LLMs may also exhibit amplified biases compared to humans \citep{li2025actions}. To counter this, we extend beyond the LLM stage by adopting a multi-perspective approach that formalizes these signals and applies solver-based verification.

Recent research in formal verification and AI fairness has shown that fairness and accountability can be treated as logical properties. Fairify \citep{biswas2023fairify} encodes externally defined fairness constraints as logic and uses SMT solvers to provide provable ethical guarantees. They employ the formal verification tool Z3 to show that a neural network satisfies these logical fairness conditions. Furthermore, \citet{singh2025position} assert that formal methods provide a reliable foundation for AI fairness and interpretability. They propose a structured specification model in which formal preconditions and postconditions define trustworthy behavior, verified through symbolic reasoning. Finally, SATLM \citep{ye2023satlm} shows that pairing LLMs with a SAT solver can yield up to 23\% higher reasoning accuracy. The model translates natural language tasks into logical constraints, then formally checks them for satisfiability. This hybrid reasoning process mirrors our architecture, which employs LLM-based moral parsing followed by solver-based coherence verification.

The specific predicate ontology adopted in this work (harm, intent, empathy, and apology for content; justification, ethical grounding, deliberation, fairness, and non-biased language for quality) was selected based on convergent coverage across established moral frameworks: Dyadic Morality \citep{schein2018theory}, Moral Foundations Theory~\citep{graham2013moral}, Image Repair Theory~\citep{benoit1997image}, Toulmin's Argumentation Model~\citep{toulmin1958uses}, and Ideal Observer Theory~\citep{smith1759moral}.
While this ontology prioritizes tractability and cross-framework grounding, we acknowledge in Section~5 that it omits dimensions such as power asymmetry, consent, and prior relational history.

Although these prior works connect formal methods and fairness in AI systems, their primary focus lies in verification of the models. In contrast, our work shifts the focus from verifying algorithmic behavior to verifying human moral reasoning.  We aggregate natural-language moral judgments by first extracting moral features using LLM-based reasoning, then formalizing these signals into logical predicates and finally applying solver-based reasoning to derive ethically coherent labels rather than relying on majority votes. Inspired by Feminist HCI principles \citep{bardzell2010feminist}, our framework considers whose reasoning and values are represented within the aggregation process. This integration of LLM reasoning, logical verification, and reflexive, human-centered ethics distinguishes our work from prior formal-methods approaches.

\section{Methodology}

\begin{figure}[h]
    \centering
    \includegraphics[width=0.6\columnwidth]{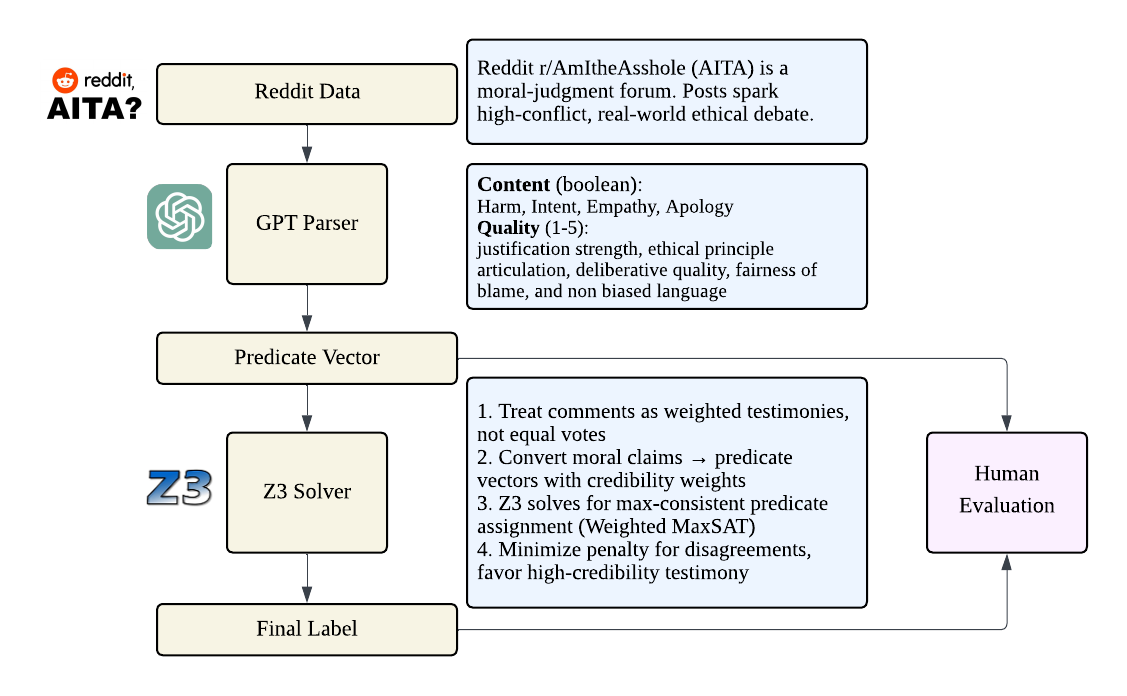}
    \caption{Overview of our reasoning-based aggregation pipeline.}
    \label{fig:reasoning_pipeline}
\end{figure}
Figure~\ref{fig:reasoning_pipeline} provides an overview of our methodology. The pipeline begins with preprocessing AITA posts and extracting top-level comments. Each comment is parsed by GPT-5.1 to generate a Content Vector (harm, intent, empathy, apology) and a Quality Vector (justification strength, fairness, ethical principles, clarity, bias). These vectors are translated into soft constraints for a Z3 MaxSAT model, where content captures moral claims and quality determines their weight. Z3 then performs a weighted optimization to produce a consistent final label. We further conduct human evaluation of both the LLM-generated vectors and the solver’s verdicts.

\subsection{Data Collection}

We use the publicly available Kaggle AITA Reddit Archive, containing over 30{,}000 complete \textit{r/AmItheAsshole} posts with associated comments~\citep{liew_reddit_kaggle}. We retain posts with at least 50 comments to ensure sufficient engagement. From these, we select 600 highly controversial posts using an entropy-based metric over label distributions (YTA, NTA, ESH, NAH).

For each post, let $p_i$ denote the proportion of comments with label $i$. We compute the label entropy:
\[
H = -\sum_i p_i \log_2(p_i).
\]

To emphasize disagreement between NTA and YTA, we weight posts as:
\[
\text{Score} = H \times \frac{\text{NTA} + \text{YTA}}{\text{Total}}.
\]

Posts with $\text{Score} > 0.9$ are selected for analysis.
\subsection{Comment Extraction and Label Processing}

For each selected post, we extract only top-level comments to capture independent moral judgments. We then identify labels (YTA, NTA, ESH, NAH) using regular expressions and compute each comment’s Reddit score as upvotes minus downvotes. Finally, we define the majority label as the label of the highest-scoring comment, reflecting the subreddit’s official verdict mechanism in which the top comment determines the final flair.

\subsection{LLM Vector Extraction}

To enable structured aggregation, we use a constrained prompting method that directs GPT-5.1 to extract two representations from each comment: a Content Vector and a Quality Vector. Both are generated in a fixed JSON schema with bounded fields and a short justification, ensuring interpretability and compatibility with downstream automated reasoning. Prior work shows that LLMs function as semantic compressors that produce stable abstractions across diverse phrasing \cite{chakraborty2025structured}, making them suitable for this vectorization task.

The \textbf{Content Vector} encodes four binary predicates: harm, intent, empathy, and apology. Harm and intent follow Dyadic Morality \cite{schein2018theory}; empathy draws on the Care/Harm foundation of Moral Foundations Theory \cite{graham2013moral}; and apology follows Benoit’s Image Repair Theory \cite{benoit1997image}. Each predicate is encoded as 0/1 with a brief justification.

The \textbf{Quality Vector} scores the epistemic reliability of a comment along five dimensions: justification strength, ethical grounding, deliberation, fairness, and unbiased language. Justification and deliberation derive from Toulmin’s Argumentation Model \cite{toulmin1958uses}, while fairness and impartiality follow Ideal Observer Theory \cite{smith1759moral}. Each feature is rated on a 1 to 5 Likert scale to support fine-grained weighting in the solver.

Together, these vectors form a compact, theory-informed abstraction that can be encoded as soft constraints for formal aggregation in Z3.

\subsection{Z3 Solver-Based Reasoning}
\begin{figure}[h]
\centering
\includegraphics[width=0.7\columnwidth]{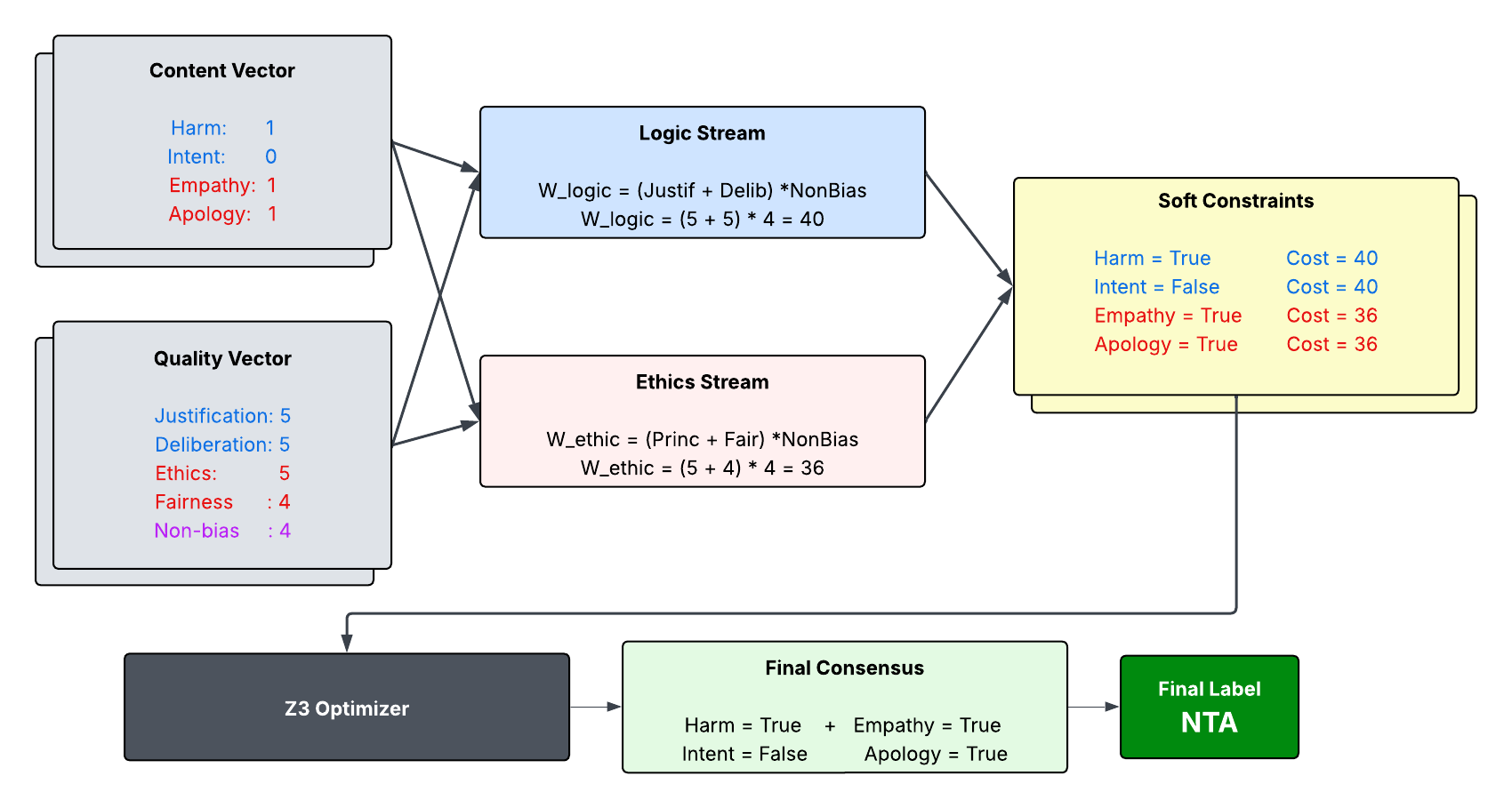}
\caption{Visual representation of the Z3 Solver-Based Reasoning Pipeline, illustrating the flow from soft-constraint assignment to verdict assignment.}
\label{fig:z3_pipeline}
\end{figure}
Following vector extraction, we employ a formal verification approach to aggregate the disparate moral judgments. Real-world morality data is rarely satisfiable in a boolean sense; conflicting comments will lead to direct logical contradictions. To address this, we replace simple aggregation with Weighted Maximum Satisfiability (MaxSAT). Rather than treating every comment as an equal vote, we treat each comment as a ``witness testimony'' with a specific credibility score. We utilize the Microsoft Z3 Theorem Prover to identify the single interpretation of events that minimizes the violation of high-quality reasoning. Figure \ref{fig:z3_pipeline} summarizes this section of the methodology.

\subsubsection{The ``Split-Stream'' Confidence Logic}
A key innovation of our pipeline is the separation of ``
reasoning quality'' into distinct epistemic categories. Drawing on Greene’s Dual Process Theory \cite{Greene2009}, which distinguishes between System 1 (Emotional/Deontological) and System 2 (Cognitive/Utilitarian) moral processing, we split the quality vector into two independent weighting streams. This ensures that a user’s competence in logical deduction does not artificially inflate their authority on emotional character judgments. This algorithm is outlined in Algorithm~\ref{alg:z3_optimization}. 

\paragraph{Stream 1: The Logical Authority ($W_{Logic}$)}
\begin{itemize}
    \item \textbf{Formula:} $W_{Logic} = (\text{Justification} + \text{Deliberation}) \times \text{Non-Bias}$
    \item \textbf{Target:} Constraints on factual predicates (Harm and Intent).
    \item \textbf{Explanation:} Establishing factual claims (e.g., ``The protagonist intentionally caused damage'') requires evidence and logical deduction. Consequently, we assign higher weight to users who demonstrate high scores in Justification and Deliberation when setting these variables.
\end{itemize}

\paragraph{Stream 2: The Ethical Authority ($W_{Ethic}$)}
\begin{itemize}
    \item \textbf{Formula:} $W_{Ethic} = (\text{Ethical Principle} + \text{Fairness}) \times \text{Non-Bias}$
    \item \textbf{Target:} Constraints on character predicates (Empathy and Apology).
    \item \textbf{Explanation:} Judging character (e.g., ``The protagonist is genuinely remorseful'') requires emotional intelligence and principled thinking. We trust users who score high on Ethical Principles and Fairness to determine these values.
\end{itemize}

The multiplicative role of Non-Bias in both streams is deliberate: rather than treating bias as a separate additive penalty, we use it as a scalar gate that proportionally suppresses the influence of prejudiced reasoning on \textit{all} predicates in that stream.
An additive formulation would allow a highly biased but otherwise well-reasoned comment to retain substantial weight; multiplication ensures that a comment scoring 1 on Non-Bias halves the effective weight of even strong logical or ethical reasoning.
The grouping of Justification and Deliberation into the Logic Stream, and Ethical Principle and Fairness into the Ethic Stream, follows directly from Greene's Dual Process Theory~\cite{Greene2009}: System~2 (deliberative/cognitive) processes are most relevant to establishing facts, while System~1 (normative/emotional) processes are most relevant to character judgments.

This split-stream architecture allows for nuanced consensus. For instance, a user might be a brilliant logician (High Stream 1) but emotionally tone-deaf (Low Stream 2). Our system incorporates their factual analysis while effectively disregarding their character judgments, creating a ``best-of-both-worlds'' consensus.

\subsubsection{The Optimization Engine}
We feed these weighted constraints into Z3 using a soft-constraint optimization strategy. Every comment generates four soft constraints (one for each predicate: Harm, Intent, Empathy, Apology). The cost of violating a constraint is equal to its calculated weight ($W_{Logic}$ or $W_{Ethic}$).

Z3 searches for a single truth assignment that minimizes the total penalty cost. If a single high-quality expert (Weight 50) disagrees with ten low-quality trolls (Weight 5 each), the solver will align with the expert because satisfying that single high-weight constraint minimizes the global cost function. This makes the system robust against mob rule and ``review bombing,'' filtering out noise and prioritizing epistemic authority.

\begin{algorithm}[h]
\caption{Consensus Extraction via Split-Stream MaxSAT}
\label{alg:z3_optimization}
\small 
\renewcommand{\baselinestretch}{0.5}\selectfont
\begin{algorithmic}[1]
\Require Set of Comments $\mathcal{C}$, where each $c \in \mathcal{C}$ has vectors $V_{content}$ and $V_{quality}$
\Ensure Consensus Booleans: $H, I, E, A$

\State \textbf{Initialize} Z3 Optimizer $\mathcal{O}$
\State \textbf{Initialize} Global Booleans $H, I, E, A$

\For{\textbf{each} comment $c$ in $\mathcal{C}$}
    \State \Comment{\textbf{Step 1: Calculate Split-Stream Weights}}
    \State $W_{logic} \gets (c_{justif} + c_{delib}) \times c_{nonbias}$
    \State $W_{ethic} \gets (c_{principle} + c_{fairness}) \times c_{nonbias}$

    \State \Comment{\textbf{Step 2: Add Logic Constraints (Head)}}
    \If{$c.harm$ is True}
        \State $\mathcal{O}$.AddSoft($H$, $W_{logic}$)
    \Else
        \State $\mathcal{O}$.AddSoft($\neg H$, $W_{logic}$)
    \EndIf
    \If{$c.intent$ is True}
        \State $\mathcal{O}$.AddSoft($I$, $W_{logic}$)
    \Else
        \State $\mathcal{O}$.AddSoft($\neg I$, $W_{logic}$)
    \EndIf

    \State \Comment{\textbf{Step 3: Add Ethic Constraints (Heart)}}
    \If{$c.empathy$ is True}
        \State $\mathcal{O}$.AddSoft($E$, $W_{ethic}$)
    \Else
        \State $\mathcal{O}$.AddSoft($\neg E$, $W_{ethic}$)
    \EndIf
    \State \Comment{(Repeat for Apology $A$ using $W_{ethic}$)}
\EndFor

\State \Comment{\textbf{Step 4: Solve}}
\If{$\mathcal{O}$.Check() == SAT}
    \State \textbf{return} $\mathcal{O}$.Model()
\Else
    \State \textbf{return} UnsatError
\EndIf

\end{algorithmic}
\end{algorithm}

\subsubsection{Logic Classification and Verdict Assignment}
Once the solver determines the consensus state of the four predicates, we apply a hierarchical decision tree to assign the final verdict. This logic is grounded in the legal principle of \textit{Mens Rea} (guilty mind) \cite{Hart1968} and the ethical framework of Restorative Justice \cite{Zehr2002}.

The decision tree (Table \ref{tab:verdict_assignment}) prioritizes actions in the following order:
\begin{itemize}
    \item \textbf{Vindication (NTA):} The system first checks for Harm. If the consensus is that no unjustified harm occurred (Harm=False), the protagonist is innocent. This captures the ``No Harm, No Foul'' principle, if the action was within the protagonist's rights, the conflict is attributed to unreasonable expectations rather than moral failure.
    \item \textbf{Malice (YTA):} If Harm exists, the system checks for Intent. Drawing from Cushman’s dual-process model \cite{Cushman2008}, we treat malice as the primary multiplier of culpability. If the harm was intentional (Intent=True), the verdict is YTA. Active aggression constitutes a ``Guilty Mind'' (\textit{Mens Rea}) \cite{Hart1968}.
    \item \textbf{The ``Accident'' Branch:} If Harm was accidental (Intent=False), the system evaluates character through the lens of the Ethics of Care \cite{Gilligan1982}.
    \begin{itemize}
        \item If the protagonist engaged in relational repair (showing Empathy or Apology), the verdict is NAH (Tragedy). This aligns with Zehr’s Restorative Justice \cite{Zehr2002}, distinguishing a ``Mistake'' (which can be repaired) from a ``Transgression.''
        \item If the protagonist refused to repair the harm, the verdict is ESH (Negligence). This defines the ``Asshole'' not by the accident, but by the Epistemic Irresponsibility \cite{Sher2009} of failing to acknowledge the harm done.
    \end{itemize}
\end{itemize}

 \begin{table}[h]
 \small
\centering
\caption{Logic Classification and Verdict Assignment}
\resizebox{\columnwidth}{!}{%
\begin{tabular}{cccccp{6cm}}
\toprule
\textbf{Harm} & \textbf{Intent} & \textbf{Empathy} & \textbf{Apology} & \textbf{Verdict} & \textbf{Logic Rule Applied} \\
\midrule
False & (Any) & (Any) & (Any) & NTA & \textbf{Rule 1: Vindication.} No unjustified harm occurred; therefore, no grievance exists. \\
True & True & (Any) & (Any) & YTA & \textbf{Rule 2: Malice.} Intentional harm is aggressive and cannot be undone by apology. \\
True & False & True & (Any) & NAH & \textbf{Rule 3A: Tragedy.} Harm was accidental, and Empathy repaired the bond. \\
True & False & False & True & NAH & \textbf{Rule 3A: Tragedy.} Harm was accidental, and Apology repaired the bond. \\
True & False & False & False & ESH & \textbf{Rule 3B: Negligence.} Harm was accidental, but the refusal to repair it is toxic. \\
\bottomrule
\end{tabular}
 \label{tab:verdict_assignment}
}
\end{table}

\subsubsection{Advantages and Limitations}
\label{sec:z3-advantages-limitations}
This aggregation methodology offers significant advantages over traditional aggregation. First, it provides noise robustness; low-quality comments are mathematically filtered out via low weights. Second, it offers explainability; unlike a black-box neural network, Z3 provides the specific variables that led to the decision.

However, we note a limitation regarding the dyadic nature of the ESH (``Everyone Sucks Here'') and NAH (``No Assholes Here'') labels. These verdicts describe an interaction where both parties share blame or innocence. Our current system is monadic, calculating the verdict based solely on the protagonist's vector (using ``Negligence'' as a proxy for ESH). To make these specific classifications more robust, future work should implement a Dual-Agent Pipeline. By extracting a parallel content vector for the antagonist, the solver could compare both vectors simultaneously, moving the system from judging individuals to judging interactions.

\section{Evaluation}

\subsection{Findings and Results}

We evaluate the pipeline in three stages: (1) human validation of the LLM-generated predicate and quality vectors, (2) analysis of Z3’s aggregation effects, and (3) human verification of solver outputs. This separation allows us to keep the human voice in morality decisions, assess the LLM as a semantic abstraction module and Z3 as a logical aggregation mechanism. We first tested 200 high-conflict posts, then scaled to 400 more, yielding 600 total posts.

\begin{figure}[h]
    \centering
    \includegraphics[width=0.4\linewidth]{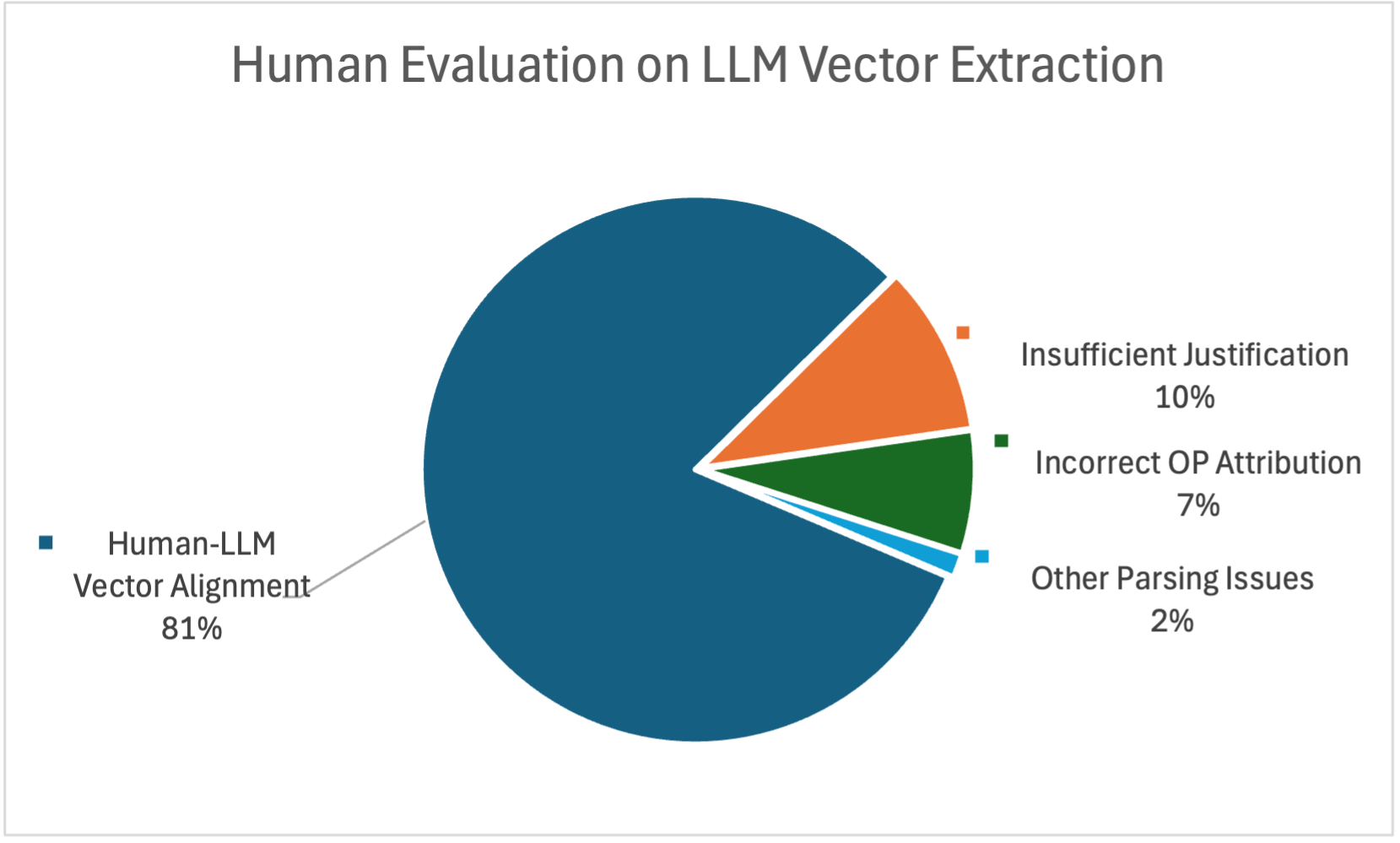}
    \caption{Pie chart summarizing results of human verification of LLM vectors}
    \label{fig:human_llm}
\end{figure}

To evaluate the correctness of LLM-derived vectors, we collected 400 human-annotated instances (20 posts × 20 comments). Reviewers judged whether the Content Vector, Quality Vector, and justification matched the evidence in the text. As shown in Figure \ref{fig:human_llm}, 81\% of comments aligned with human interpretation. The remaining cases showed structured and predictable failure types: 10\% Insufficient Justification, 7\% Incorrect OP Attribution, and 2\% other parsing issues. These patterns indicate that GPT-5.1 functions as a stable semantic parser suitable for downstream formal reasoning.

We next measured Z3’s impact on final labels. Across 600 posts, the solver changed Reddit’s majority verdict in 372 cases (62\%). This reflects how vote-based aggregation can overweight low-quality arguments, whereas the solver enforces a uniform logical structure and weights testimonies by epistemic quality. The frequent divergence from the top-upvoted comment suggests that MaxSAT aggregation identifies a more coherent global assignment than majority voting.

Finally, to test whether solver-derived decisions align with independent human reasoning, we conducted human evaluation, with three independent reviewers, on 50 high-conflict posts, blind to both Reddit’s majority label and the solver’s internal predicate assignments. The resulting alignment of 86\% indicates that although the solver frequently corrects the crowd-assigned label, its decisions remain consistent with independent human interpretations. We note that these evaluators do not constitute a bias-free normative ground truth; like all human raters, they bring their own moral sensibilities. Rather, this result is interpreted as a consistency check: the solver's outputs are not idiosyncratic artifacts of the formalism, but align with a sample of human reasoning.

An example of a changed decision can be seen in Appendix \ref{app:example_changed_decision}. Overall, the evaluations demonstrate that the LLM–solver pipeline produces stable abstractions, robust logical aggregation, and human-aligned outcomes, supporting its use as an interpretable automated reasoning framework for multi-annotator judgment tasks.

\subsection{Discussion and Implications for Automated Reasoning and AI Ethics}

Our results show that the LLM-Z3 pipeline frequently overturns Reddit’s majority verdicts, most commonly shifting NTA to YTA. Unlike upvote-based aggregation, which implicitly treats upvotes as a signal of correctness, we aggregate all comments as weighted soft constraints and use MaxSAT to find the predicate assignment that minimizes global inconsistency. As a result, lenient crowd labels are revised when they conflict with higher-weight constraints. 

We identify two primary mechanisms driving this shift toward stricter accountability. The first driver is narrative incompleteness. Consistent with Issendai’s “missing missing reasons” framework \cite{issendai_missing_missing_reasons}, OPs often omit context to construct a favorable self-narrative. Higher-quality commenters are more likely to flag these gaps, suggest alternative interpretations, or scrutinize inconsistencies. Weighting those critiques more heavily prevents the solver from treating the OP’s framing as ground truth, correcting for the information asymmetry that sways popular voting.

\begin{figure}[h]
    \centering
    \includegraphics[width=0.6\linewidth]{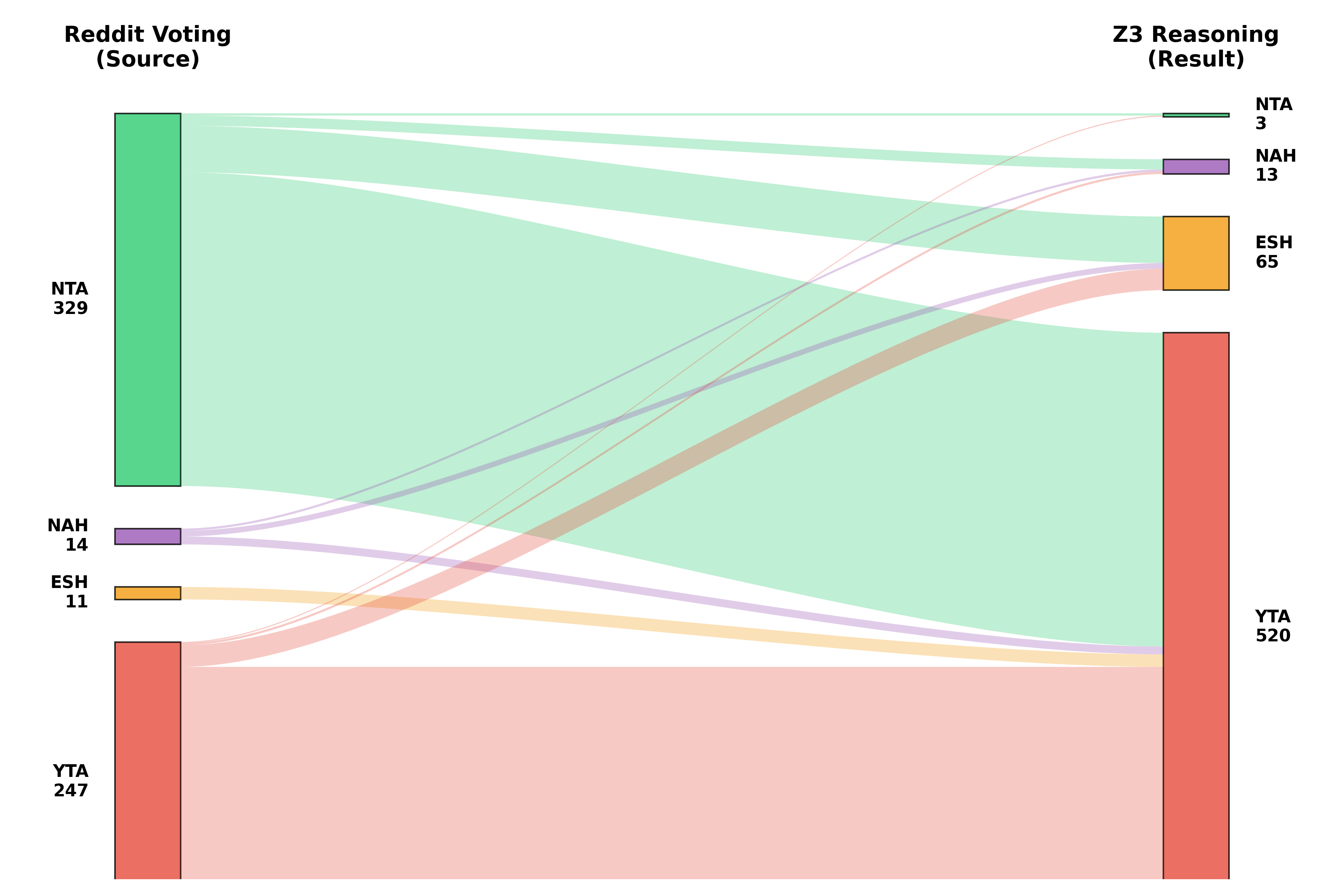}
    \caption{Sankey diagram of changed final decisions from Reddit to proposed pipeline}
    \label{fig:flow_diag}
\end{figure}

Second, the pipeline effectively disrupts social conformity bias. Majority verdicts are frequently shaped by ``echo chamber'' dynamics, where early validation can anchor later comments and amplify the OP’s framing. Because our architecture separates argumentation quality from popularity, the final verdict is less sensitive to herd dynamics and instead reflects the most internally consistent set of ethical predicates, showing immunity to the ``hivemind" effect. Figure~\ref{fig:flow_diag} visualizes label transitions, dominated by NTA$\rightarrow$YTA with smaller shifts into ESH, matching the overall strictness increase under constraint-based aggregation. 

Finally, Appendix~\ref{app:conf_mat} provides the full confusion matrix. The concentration of mass along the NTA $\rightarrow$ YTA and YTA $\rightarrow$ YTA entries highlights that the solver’s main departures are toward higher accountability rather than random relabeling. These results underscore the application of reasoning-based pipelines to transform content moderation in high-conflict online communities. Instead of relying on sentiment or upvotes, the pipeline evaluates the logical structure and ethical framing of comments, helping surface constructive moral reasoning and downrank popular but harmful rhetoric (including “smart trolls” and mob harassment).

The findings illustrate the complementary strengths of LLMs and SAT solvers. LLMs provide rich semantic structure, while Z3 enforces global coherence by requiring that all predicates jointly satisfy the weighted constraint set. In this hybrid architecture, the solver’s role is not to judge morality but to perform consistency management over human-provided rationales. Overall, these results suggest that a neuro-symbolic aggregation framework can improve transparency and robustness in collective reasoning when combined with appropriate human oversight. Rather than replacing human judgment, automated reasoning can help surface hidden assumptions, elevate minority perspectives, and support more coherent deliberation in settings with incomplete or contested narratives. 
Beyond content moderation on AITA, the reasoning-based aggregation paradigm presented here generalizes naturally to other high-conflict judgment domains.
In \textit{political deliberation}, structured predicate extraction could surface principled minority arguments in policy debates, providing a counterweight to majoritarian dynamics in participatory platforms.
In \textit{legal contexts}, the split-stream architecture maps onto existing jurisprudential distinctions---such as the separation of factual findings from character assessments in sentencing---suggesting a path toward more interpretable automated legal reasoning aids.
More broadly, any decision task where reasoning quality systematically varies across deciders (e.g., misinformation labeling, hate speech detection, medical ethics review) is a candidate for this framework, with the caveat that predicate ontologies would need to be re-grounded in domain-appropriate moral or epistemic frameworks.

\section{Limitations and Future Work}

While the proposed pipeline provides a structured mechanism for aggregating moral judgments, some limitations point to important directions for future work. Because the solver optimizes only over extracted predicates, errors or cultural biases introduced by the single LLM front-end may propagate into the reasoning layer; Z3 can reweight constraints but cannot repair incorrect abstractions, recover omitted context, or reconcile competing moral frameworks. Future iterations could therefore incorporate multiple LLMs, lightweight role-tracking and coreference validation, and automated consistency checks that compare extracted predicates against quoted evidence. Evaluation should also be expanded beyond a small, relatively homogeneous annotator pool with single reviews per post, enabling measurement of inter-annotator agreement and systematic disagreement across diverse perspectives. Finally, the current predicate ontology, limited to four content predicates and five quality dimensions, prioritizes tractability at the expense of social nuance, omitting factors such as power asymmetry, prior history, consent, and proportionality, and collapsing epistemic authority into a single scalar. Extending the framework toward richer, potentially multi-agent predicate representations and community-specific weighting schemes within a shared logical backbone represents a promising path forward.

\section{Conclusion}
This study presents a neuro-symbolic aggregation framework for aggregating moral judgments that replaces majority consensus with Weighted Maximum Satisfiability (MaxSAT). By coupling LLMs for semantic abstraction with the Z3 solver for aggregation, the pipeline treats community comments as weighted testimony, utilizing a novel ``split-stream" architecture to decouple factual validity from normative character judgments. This formal approach enforces logical constraints grounded in reasoning, effectively filtering out inconsistent or socially conformist decisions that plague standard voting mechanisms. Empirically, this optimization strategy produced a measurable ``strictness shift" across 600 high-conflict scenarios, overturning 62\% of popularity-based verdicts while achieving 86\% agreement with independent human evaluation. These results show that integrating neural perception with symbolic constraint solving provides a scalable blueprint for moderation platforms to automate principled accountability grounded in ethical logic while avoiding toxic majoritarianism.


\printbibliography[heading=subbibintoc]

@inproceedings{z3solver,
author = {De Moura, Leonardo and Bj\o{}rner, Nikolaj},
title = {Z3: an efficient SMT solver},
year = {2008},
isbn = {3540787992},
publisher = {Springer-Verlag},
address = {Berlin, Heidelberg},
booktitle = {Proceedings of the Theory and Practice of Software, 14th International Conference on Tools and Algorithms for the Construction and Analysis of Systems},
pages = {337–340},
numpages = {4},
location = {Budapest, Hungary},
series = {TACAS'08/ETAPS'08}
}

@article{schein2018theory,
  title={The theory of dyadic morality: Reinventing moral judgment by redefining harm},
  author={Schein, Chelsea and Gray, Kurt},
  journal={Personality and social psychology review},
  volume={22},
  number={1},
  pages={32--70},
  year={2018},
  publisher={Sage Publications Sage CA: Los Angeles, CA}
}

@incollection{graham2013moral,
  title={Moral foundations theory: The pragmatic validity of moral pluralism},
  author={Graham, Jesse and Haidt, Jonathan and Koleva, Sena and Motyl, Matt and Iyer, Ravi and Wojcik, Sean P and Ditto, Peter H},
  booktitle={Advances in experimental social psychology},
  volume={47},
  pages={55--130},
  year={2013},
  publisher={Elsevier}
}

@article{benoit1997image,
  title={Image repair discourse and crisis communication},
  author={Benoit, William L},
  journal={Public relations review},
  volume={23},
  number={2},
  pages={177--186},
  year={1997},
  publisher={Elsevier}
}

@book{smith1759moral,
  author    = {Adam Smith},
  title     = {The Theory of Moral Sentiments},
  year      = {1759},
  publisher = {A. Millar; and A. Kincaid and J. Bell, Edinburgh},
  address   = {London / Edinburgh}
}

@book{toulmin1958uses,
  author    = {Stephen E. Toulmin},
  title     = {The Uses of Argument},
  year      = {1958},
  publisher = {Cambridge University Press},
  address   = {Cambridge}
}

@misc{issendai_missing_missing_reasons,
  author       = {Issendai},
  title        = {The Missing Missing Reasons},
  howpublished = {\url{https://www.issendai.com/psychology/estrangement/missing-missing-reasons.html}}
}

@book{Hart1968,
  author = {Hart, H. L. A.},
  title = {Punishment and Responsibility: Essays in the Philosophy of Law},
  publisher = {Oxford University Press},
  year = {1968}
}

@book{Zehr2002,
  author = {Zehr, H.},
  title = {The Little Book of Restorative Justice},
  publisher = {Good Books},
  year = {2002}
}

@article{Cushman2008,
  author = {Cushman, F.},
  title = {Crime and Punishment: Distinguishing the Roles of Causal and Intentional Analyses in Moral Judgment},
  journal = {Cognition},
  volume = {108},
  number = {2},
  pages = {353--380},
  year = {2008}
}

@book{Gilligan1982,
  author = {Gilligan, C.},
  title = {In a Different Voice: Psychological Theory and Women’s Development},
  publisher = {Harvard University Press},
  year = {1982}
}

@book{Sher2009,
  author = {Sher, G.},
  title = {Who Knew? Responsibility Without Awareness},
  publisher = {Oxford University Press},
  year = {2009}
}

@article{Greene2009,
  author = {Greene, Joshua D.},
  title = {Dual-process morality and the personal/impersonal distinction: A reply to McGuire, Langdon, Coltheart, and Mackenzie},
  journal = {Journal of Experimental Social Psychology},
  volume = {45},
  number = {3},
  pages = {581--584},
  year = {2009}
}

@inproceedings{gordon2022jurylearning,
  author    = {Gordon, Mitchell L. and Lam, Michael S. and Park, Juho S. and Patel, Kayur and Hancock, Jeff T. and Hashimoto, Tatsunori and Bernstein, Michael S.},
  title     = {Jury Learning: Integrating Dissenting Voices into Machine Learning Models},
  booktitle = {CHI Conference on Human Factors in Computing Systems (CHI '22)},
}

@misc{liew_reddit_kaggle,
  author       = {Liew, Jian Loong},
  title        = {Reddit Dataset},
  howpublished = {Kaggle dataset},
  url          = {https://www.kaggle.com/datasets/jianloongliew/reddit},
  year         = {2022}

}

@inproceedings{zhang2023magi,
  author    = {Zhang, Y. and Gu, S. and Gao, Y. and Pan, B. and Yang, X. and Zhao, L.},
  title     = {MAGI: Multi-Annotated Explanation-Guided Learning},
  booktitle = {Proceedings of the IEEE/CVF International Conference on Computer Vision (ICCV)},
  year      = {2023}
}

@misc{bsher2023aita,
      title={AITA Generating Moral Judgements of the Crowd with Reasoning}, 
      author={Osama Bsher and Ameer Sabri},
      year={2023},
      eprint={2310.18336},
      archivePrefix={arXiv},
      primaryClass={cs.CL},
      url={https://arxiv.org/abs/2310.18336}, 
}

@inproceedings{lourie2021scruples,
  author    = {Lourie, Nicholas and Le Bras, Ronan and Choi, Yejin},
  title     = {SCRUPLES: A Corpus of Community Ethical Judgments on 32,000 Real-life Anecdotes},
  booktitle = {Proceedings of the Thirty-Fifth AAAI Conference on Artificial Intelligence (AAAI-21)},
  year      = {2021},
  pages     = {13470--13479}
}

@article{chakraborty2025structured,
  author       = {Chakraborty, Mohna and Wang, Lu and Jurgens, David},
  title        = {Structured Moral Reasoning in Language Models: A Value-Grounded Evaluation Framework},
  year         = {2025},
  eprint       = {2506.14948},
  archivePrefix= {arXiv},
  primaryClass = {cs.HC},
  doi          = {10.48550/arXiv.2506.14948},
  note         = {arXiv:2506.14948 [cs.HC]}
}

@article{li2025actions,
  author       = {Li, Yuxuan and Shirado, Hirokazu and Das, Sauvik},
  title        = {Actions Speak Louder than Words: Agent Decisions Reveal Implicit Biases in Language Models},
  year         = {2025},
  doi          = {10.48550/arXiv.2501.17420},

}

@inproceedings{biswas2023fairify,
  author    = {Biswas, Sumon and Rajan, Hridesh},
  title     = {Fairify: Fairness Verification of Neural Networks},
  booktitle = {Proceedings of the 45th IEEE/ACM International Conference on Software Engineering (ICSE 2023)},
  year      = {2023},
  pages     = {1546--1558},
  publisher = {IEEE}
}

@inproceedings{singh2025position,
  author    = {Singh, G. and Chawla, D.},
  title     = {Position: Formal Methods are the Principled Foundation of Safe AI},
  booktitle = {Workshop on Technical AI Governance (TAIG) at ICML 2025},
  year      = {2025},
  address   = {Vancouver, Canada}
}

@inproceedings{ye2023satlm,
  author    = {Ye, X. and Chen, Q. and Dillig, I. and Durrett, G.},
  title     = {SATLM: Satisfiability-Aided Language Models Using Declarative Prompting},
  booktitle = {Proceedings of the 37th Conference on Neural Information Processing Systems (NeurIPS 2023)},
  year      = {2023}
}

@inproceedings{bardzell2010feminist,
  author    = {Bardzell, S.},
  title     = {Feminist HCI: Taking Stock and Outlining an Agenda for Design},
  booktitle = {Proceedings of the SIGCHI Conference on Human Factors in Computing Systems},
  year      = {2010},
  pages     = {1301--1310}
}

@misc{aita2022example,
  author       = {{Anonymous Reddit User}},
  title        = {Am I the Asshole?},
  howpublished = {\url{https://www.reddit.com/r/AmItheAsshole/}},
  year         = {2022},
  note         = {Accessed for research purposes; original post anonymized}
}

\appendix
\section{}

\label{app:additional}

\subsection{Example of a changed decision}
    \label{app:example_changed_decision}

\begin{figure}[H]
     \centering
     \includegraphics[width=0.5\linewidth]{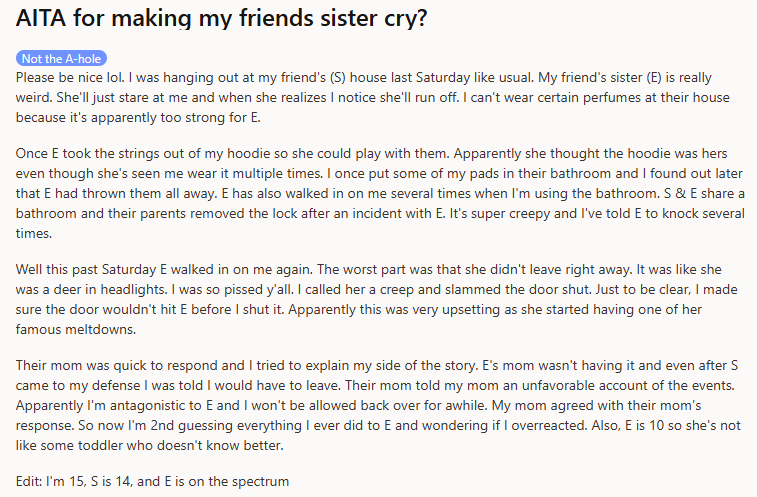}
     \caption{Screenshot of an original post from the r/AmItheAsshole subreddit \citep{aita2022example}.}
    \label{fig:example_changed_decision}
 \end{figure}

Figure~\ref{fig:example_changed_decision} shows a post labeled NTA by majority-voting that our pipeline reclassifies as YTA under weighted MaxSAT aggregation. While many comments excuse the OP by focusing on the sister’s intrusive behavior, several high-quality critiques highlight an overlooked detail: the confronted individual is described as autistic, and the OP’s response (“creep” + door-slam) constitutes avoidable harm with insufficient empathy and repair.

Because the solver upweights better-justified, less biased reasoning, these critiques dominate the constraint set and shift the final label from the most popular reaction to the most internally consistent ethical account. This example illustrates how solver-based aggregation can preserve well-supported minority arguments within a thread that would otherwise be discounted under frequency- or upvote-based labeling.



\subsection{Confusion matrix of pipeline results}
    \label{app:conf_mat}

\begin{figure}[H]
    \centering
    \includegraphics[width=0.5\linewidth]{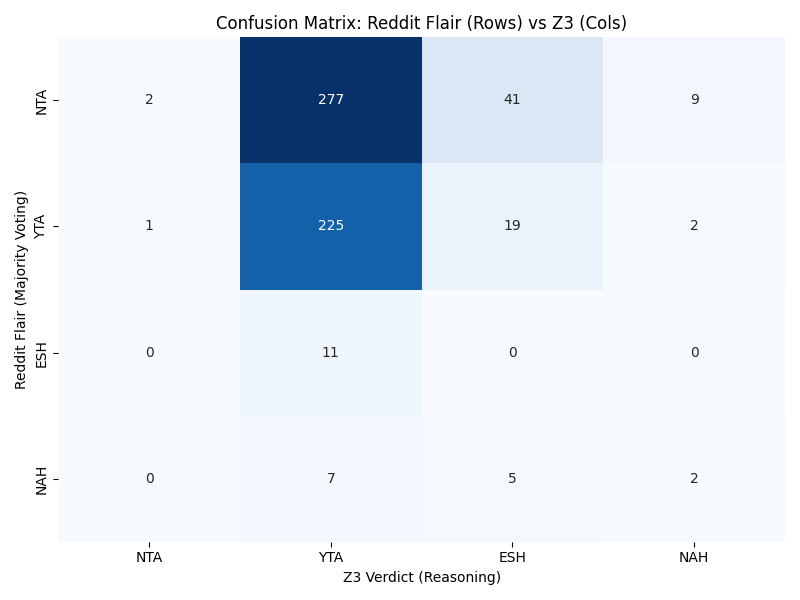}
    \caption{Confusion matrix of final decisions from Reddit majority vote and the pipeline}
    \label{fig:conf_mat}
\end{figure}


\subsection*{Appendix B. Reproducibility Details}

\subsubsection*{B.1. LLM Configuration}
All vector extraction was performed using GPT-5.1 via the OpenAI API.
We used a temperature of \textbf{0.0} to maximize determinism and reproducibility of structured outputs.
Maximum output tokens were set to 1000 per batch call.
Each API call processed one post and its associated comments in a single prompt to preserve shared context across the comment set.

\subsubsection*{B.2. Output Schema}
The LLM was instructed to return a strictly valid JSON object with the following schema:

\begin{verbatim}
{
  "analyses": [
    {
      "comment_id": "<string>",
      "comment_content_vector": [harm, intent, empathy, apology],
      "comment_quality_vector": [justif, ethic, delib, fairness, nonbias],
      "reasoning": "<max two sentences>"
    }
  ]
}
\end{verbatim}

\noindent where \texttt{comment\_content\_vector} entries are integers in $\{0,1\}$ and \texttt{comment\_quality\_vector} entries are integers in $\{1,\ldots,5\}$.
The prompt explicitly prohibited trailing commas and any non-JSON preamble.

\subsubsection*{B.3. System Prompt}
The system prompt provided to the LLM is reproduced below. Dimension definitions were grounded in the moral frameworks cited in Section~3.3.

\begin{quote}
\small
\textit{``You are a fair and neutral comment evaluator. Your task is to evaluate comments on a Reddit `Am I the Asshole' (AITA) post. [\ldots] Each dimension is grounded in explicit moral theories to maintain interpretive consistency: intentionality follows Dyadic Morality (Schein \& Gray, 2018); harm follows Dyadic Morality; empathy draws on the Care/Harm foundation of Moral Foundations Theory (Graham et al., 2013); apology follows Image Repair Theory (Benoit, 1997). Quality dimensions draw on Toulmin's Argumentation Model (1958) and Ideal Observer Theory (Smith, 1759). Your output MUST be valid JSON.''}
\end{quote}

\noindent The full prompt text, including all dimension definitions, is available in the code repository linked in footnote~1.

\subsubsection*{B.4. Z3 Solver Objective}
The solver objective minimizes total soft-constraint violation cost.
For each comment $c$ and each predicate $p \in \{\text{Harm, Intent, Empathy, Apology}\}$, a soft constraint is added with cost $W_{\text{logic}}$ (for Harm, Intent) or $W_{\text{ethic}}$ (for Empathy, Apology), as defined in Section~3.4.1.
Z3's \texttt{Optimize} module (Weighted MaxSAT mode) is used; hard constraints are not added.
The full solver implementation is provided in the code repository.

\end{document}